\documentclass[11pt]{article}

\usepackage[final]{acl}

\usepackage{times}
\usepackage{latexsym}

\usepackage[T1]{fontenc}

\usepackage[utf8]{inputenc}

\usepackage{microtype}

\usepackage{inconsolata}

\usepackage{graphicx}
\usepackage{mathtools}
\usepackage{amssymb}
\usepackage{arydshln}
\usepackage{tcolorbox}

\title{Generating Storytelling Images with Rich Chains-of-Reasoning}

\author{
    Xiujie Song$^1$, Qi Jia$^2$, Shota Watanabe$^{1}$, Xiaoyi Pang$^{1}$, \\
    \textbf{Ruijie Chen$^3$, Mengyue Wu$^{1,}$\footnotemark[1], Kenny Q. Zhu$^{4,}$\thanks{Corresponding authors.}} \\
    $^1$X-LANCE Lab, School of Computer Science, Shanghai Jiao Tong University, China\\
    $^2$Shanghai Artificial Intelligence Laboratory, China\\
    $^3$East China University of Technology, China\\
    $^4$University of Texas at Arlington, USA\\
    \texttt{\{xiujiesong, soawb71, fointpang, mengyuewu\}@sjtu.edu.cn, } \\
    \texttt{jiaqi@pjlab.org.cn, 2023213701@ecut.edu.cn, kenny.zhu@uta.edu}
}

\begin{document}
\maketitle

\begin{abstract}
    A single image can convey a compelling story through logically connected visual clues, forming Chains-of-Reasoning (CoRs). 
    We define these semantically rich images as Storytelling Images. 
    By conveying multi-layered information that inspires active interpretation, these images enable a wide range of applications, such as illustration and cognitive screening. 
    Despite their potential, such images are scarce and complex to create. 
    To address this, we introduce the \textbf{Storytelling Image Generation} task and propose StorytellingPainter, a two-stage pipeline combining the reasoning of Large Language Models (LLMs) with Text-to-Image (T2I) synthesis. 
    We also develop a dedicated evaluation framework assessing semantic complexity, diversity, and text-image alignment. 
    Furthermore, given the critical role of story generation in the task, we introduce lightweight Mini-Storytellers to bridge the performance gap between small-scale and proprietary LLMs. 
    Experimental results demonstrate the feasibility of our approaches\footnote{Code available at: \url{https://github.com/xiujiesong/StorytellingImageGeneration}}. 
\end{abstract}

\section{Introduction}

\begin{center}
    \vspace*{1em}
    \textit{``A picture is worth a thousand words.''}
    \vspace*{1em}
\end{center}

Is it possible for a single image to tell a logically coherent and semantically rich story on its own? 
The answer is yes, and we define this kind of image as \textit{Storytelling Image} in this paper, which typically conveys a story through rich visual clues and logical connections among them.

\begin{figure}[h!]
    \centering
    \includegraphics[width=0.48\textwidth]{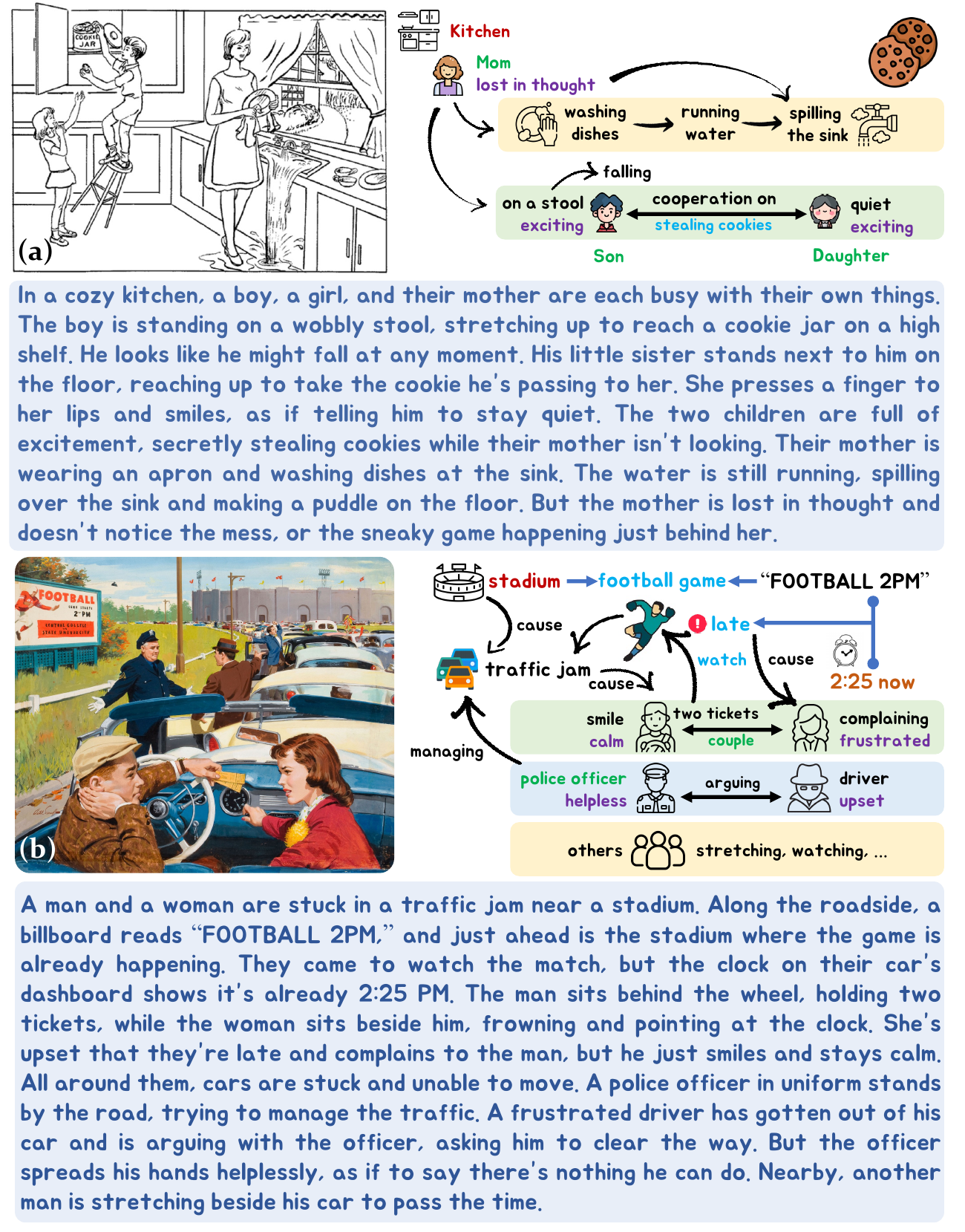}
    \caption{Storytelling Images with rich CoRs. Each case consists of the image, a graph formed by the connected CoRs, and a description. (a) Cookie Theft, a widely used image in cognitive and linguistic assessments. (b) Frustration, a magazine cover illustration. 
    }
    \label{fig:examples}
\end{figure}

The Cookie Theft picture (Figure~\ref{fig:examples} (a)) from Boston Diagnostic Aphasia Examination (BDAE)~\cite{goodglass2001-ej} is a typical Storytelling Image with rich semantics. 
It tells the story of two children sneaking cookies from the cupboard while their mother is doing the dishes. 
It is precisely because of the rich semantics embedded in this image that it has been widely used in picture description tasks to evaluate human cognitive and language abilities for over 50 years. 
This picture is still widely used, but is now considered outdated and does not adapt well across different cultures~\cite{berube2019stealing, rethinkingct}.  
To address these issues, it is often modified or replaced to suit different application scenarios~\cite{HUSSEIN2015152, DOMINGUEZ2006476, Oh2012ValidityAR}. 
In addition to human cognitive assessment, this type of image can also be used to evaluate the cognitive reasoning abilities of Large Vision-Language Models (LVLMs)~\cite{song-etal-2025-cognitive}. 
Beyond the evaluative function, these images often appear as illustrations in magazines and books, created by accomplished artists~\cite{claridge2001norman}. 
This indicates that such Storytelling Images are rare yet in high demand, and that generating them remains challenging even for experts. 
Therefore, exploring how to automatically create such images holds substantial scientific and practical value.

Storytelling Images convey compelling narratives through rich visual clues that are semantically and logically interconnected. We define a sequence that links these visual clues with their corresponding inferential conclusions as a Chain-of-Reasoning (CoR),  which enables the image to convey information across multiple semantic dimensions to the viewers. 
These semantic dimensions include characters and their relationships, events and their causal connections, as well as the mental states of the characters, etc.~\cite{describe-ctp}.
In other words, these images are characterized by high semantic density, which makes their creation particularly challenging. 
For instance, in Figure~\ref{fig:examples} (b), visual clues such as \textit{the “FOOTBALL 2 PM” sign, the stadium, the police officer, the two tickets, the 2:25 clock, and the woman’s expression} form a CoR leading to the conclusion that [the couple is stuck in traffic and will likely miss the game].

Although image generation is a classic problem, no existing work targets such semantically complex images. 
Story Visualization~\cite{8953914} centers on generating a sequence of images to narrate, yet remains distinct from generating a single Storytelling Image that demands higher semantic density. 
Previous approaches have struggled to generate such images because the task demands not only advanced reasoning but also high-quality image synthesis capabilities, making it particularly challenging. 
Thus, motivated by the inherent appeal, wide applicability, and extreme scarcity of such images, we propose the Storytelling Image Generation task to fill this research gap.

Human creation of these images generally follows a two-stage process: story design and image rendering. 
For example, images like Cookie Theft used for cognitive assessment are typically drawn by an artist based on the specifications provided by professionals~\cite{nicholas1993system}. 
To simulate this process, we propose a two-stage pipeline, StorytellingPainter, which consists of a Large Language Model (LLM) as Storyteller model and a Text-to-Image (T2I) model as Painter model. 
To comprehensively evaluate the task and pipeline, we also propose an evaluation framework consisting of three evaluators: a Semantic Complexity Evaluator for the semantic complexity of generated stories and images; a KNN-based Diversity Evaluator for story diversity; and a Story-Image Alignment Evaluator for the alignment between stories and generated images. 
Given that the story serves as the foundation of a Storytelling Image, the observed performance gap between smaller open-source and proprietary LLMs is a critical bottleneck. 
To bridge this, we explore strategies to train computationally efficient Mini-Storyteller models.

To sum up, our contributions are as follows:
\begin{itemize}
    \item 
    We are the first to formalize the Storytelling Image Generation task, which focuses on generating a single image to tell a semantically rich story with various CoRs. 
    \item We propose StorytellingPainter, a pipeline achieving state-of-the-art results in Storytelling Image Generation. Complementing this pipeline, we introduce a robust evaluation framework featuring three key evaluators. We validate our approach through extensive experiments across diverse configurations. 
    \item We explore different training strategies to enhance the storytelling ability of lightweight open-source LLMs. The resulting Mini-Storyteller models show substantial improvements in story generation quality. 
\end{itemize}
\section{StorytellingPainter} 
\label{sec:pipeline}

This section introduces the Storytelling Image Generation task, the StorytellingPainter pipeline, and the evaluation framework, concluding with the experimental analysis.

\begin{figure*}[h!]
    \centering
    \includegraphics[width=0.86\textwidth]{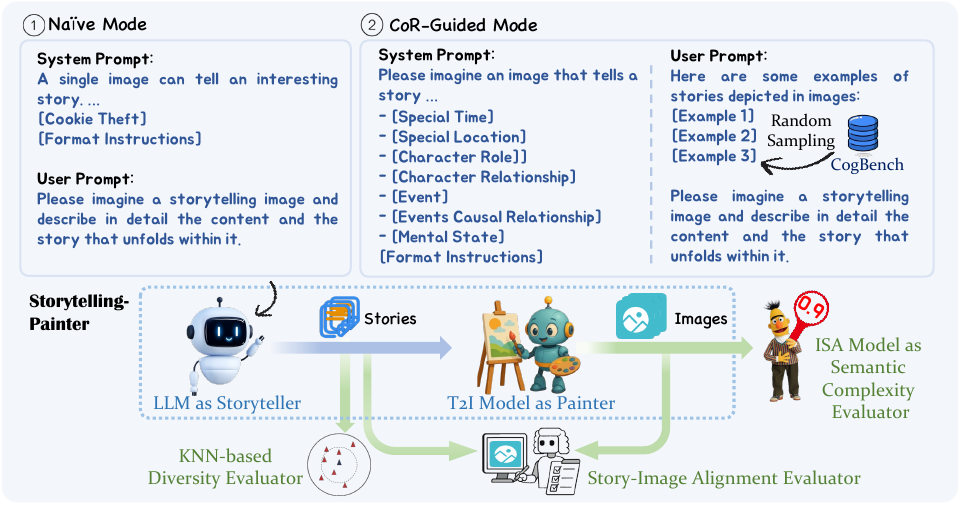}
    \caption{Our proposed StorytellingPainter pipeline and dedicated evaluators. 
    }
    \label{fig:sig_pipeline}
\end{figure*}

\subsection{Task Definition}

The goal of \textbf{Storytelling Image Generation} task is to leverage the creative capabilities of generative AI models to produce diverse semantically rich Storytelling Images with minimal human effort. 

A semantically rich \textbf{Storytelling Image} is defined as one that embeds complex \textbf{Chains-of-Reasoning (CoRs)}~\cite{song-etal-2025-cognitive}, which logically connect visual clues to form a narrative.

\subsubsection*{Definition of Chain-of-Reasoning}

Let $C = \{c_1, c_2, \dots, c_n\}$ be a set of distinct visual clues identified within the image. 
Let $K$ be an inferred conclusion. 
The CoR is defined by the logical structure where the \textit{conjunction} of these clues implies the conclusion $K$:

{\small
\begin{equation}
\text{CoR} \coloneqq (c_1 \land c_2 \land \dots \land c_n) \implies K
\end{equation}
}

Here, $\land$ denotes logical conjunction, and $\implies$ denotes logical implication. 
This structure asserts that the combined presence of all visual clues $\{c_i\}$ logically supports the inferred conclusion $K$.

\subsubsection*{Definition of Storytelling Image Generation}
\label{sec:task_def}

Given $N$ instructions $I=\{i_1, \dots, i_N\}$ where elements are not necessarily distinct, the task is to first create $N$ stories $S=\{s_1, \dots, s_N\}$ and then generate $N$ corresponding Storytelling Images $Y=\{y_1, \dots, y_N\}$.

Each story $s_i \in S$ should describe a dramatic, single-moment scene that can be depicted in one image, rather than a long story with a time sequence or dialogue. 
It should also be semantically distinct from all other stories in the set as much as possible.
Each generated image $y_i \in Y$ is expected to independently and accurately convey the story $s_i$.

\subsection{StorytellingPainter Pipeline} 
\label{sec:sig_method}

When human experts create such images, they normally first envision a story and then bring that imagined scene to life on the canvas with their brush. 
Inspired by this process, we propose a two-stage StorytellingPainter pipeline, as shown in Figure~\ref{fig:sig_pipeline}. 
In the first stage, we leverage the imagination and reasoning abilities of LLMs to generate a story, referring to this component as the Storyteller model. 
The key difference between this type of story and typical narratives lies in the temporal structure. The stories we generate here are anchored in a single moment in time, meaning each character can perform only one action. In contrast, typical stories unfold over a sequence of events, allowing characters to perform multiple actions in succession. 
In the second stage, we leverage the image generation capabilities of T2I models to create images that depict the generated stories, referring to this component as the Painter model.

For the Storyteller model, we explore two different prompting modes, as illustrated in Figure~\ref{fig:sig_pipeline}.

\paragraph{Naive Mode}
We use a relatively simple prompt for the Storyteller. 
The prompt specifies requirements for imagining a story that can be told by an image and provides the Cookie Theft story as an example and some format instructions. 
The format instructions specify our requirements for the structure of the generated stories. For example, we prefer the stories to be concise rather than lengthy, and they should be imaginative interpretations of a single image rather than traditional narratives with sequences of events or dialogues.

\paragraph{CoR-Guided Mode}
\label{sec:corg}

The CoRs that constitute a Storytelling Image often involve multiple semantic dimensions. 
By analyzing the Cookie Theft and other Storytelling Images, we identify seven core dimensions that capture the essence of Storytelling Images.
We incorporate these dimensions as prior knowledge to guide the Storyteller model in generating more coherent and semantically rich stories.

\begin{itemize}
    \item \textbf{[Time]}: The specific time when the story takes place, such as Christmas or 3:00 AM.
    \item \textbf{[Location]}: The specific setting of the story, such as a grocery store or a baseball field.
    \item \textbf{[Character Role]}: The roles of the characters, such as a teacher or a firefighter.
    \item \textbf{[Character Relationship]}: The relationships between characters, such as a mother and son. 
    \item \textbf{[Event]}: The key events that occur in the story. For example, two boys stealing cookies.
    \item \textbf{[Event Causal Relationship]}: The causal relationships between events. For example, a boy is being scolded because he is doodling. 
    \item \textbf{[Mental State]}: The emotional or psychological states of the characters, such as happiness. 
\end{itemize}

Leveraging CogBench's~\cite{song-etal-2025-cognitive} rich collection of Storytelling Images and descriptions, we propose a Dynamic In-Context Learning (Dyn-ICL) strategy to enhancing diversity. 
It randomly samples three descriptions from CogBench as in-context examples for each Storyteller prompt.

\subsection{Evaluators}

Based on the task objectives in Section~\ref{sec:task_def}, we designed a dedicated evaluation framework for the task and pipeline. 
Specifically, the framework comprises three evaluators: a Semantic Complexity Evaluator to evaluate the semantic complexity of generated stories and images, a KNN-based Diversity Evaluator to evaluate the diversity of stories and a Story-Image Alignment Evaluator to measure the semantic gap between stories and images.

\subsubsection{Semantic Complexity Evaluator}
The Image Semantic Assessment (ISA) task~\cite{Song_Pang_Tang_Wu_Zhu_2025} aligns well with the objective of our task, as it aims to automatically evaluate the storytelling ability of an image by defining a Semantic Complexity Score. 
Therefore, we adopt the ISA model, CoT VLISA (BERT), as a key Evaluator for our pipeline to automatically calculate Semantic Complexity Score. 
CoT VLISA (BERT) first extracts features from the image in text form using GPT-4o~\cite{Achiam2023GPT4TR} and then computes the Semantic Score based on the text features with a fine-tuned BERT~\cite{devlin-etal-2019-bert}. 
The extracted text features include the seven dimensions of visual clues mentioned in the CoR-Guided Mode in Section~\ref{sec:sig_method}. 
With Semantic Score, the effectiveness of Storytellers and Painters can be assessed.

\subsubsection{KNN-based Diversity Evaluator}

Inspired by studies on dataset diversity~\cite{stasaski-hearst-2022-semantic,stasaski-etal-2020-diverse,yang-etal-2025-measuring}, we adopt distance-based semantic diversity to measure the diversity of our generated stories. 

To calculate Diversity Score, we first address the potential bias introduced by the varying story lengths produced by different models. 
To ensure a fair comparison focused on core narrative components, we employ a custom-designed summarizer (GPT-4.1-mini), which processes each generated story to extract a relatively standardized, structured representation comprising four CoR elements: [Time], [Location], [Character], and [Event]. 
Subsequently, the extracted information is input into Qwen3-Embedding-0.6B~\cite{qwen2025qwen25technicalreport} to yield the corresponding embedding $e$. 
The Diversity Score is defined as the average KNN cosine distance across all embeddings.

{\small
\begin{equation}
\text{Diversity Score} = \frac{1}{N} \sum_{i=1}^{N} \frac{1}{K} \sum_{j \in \text{KNN}(i)} (1 - \cos(e_i, e_j)),
\end{equation}
}

where $N$ is the number of generated stories, $K$ is the number of nearest neighbors, and $e_i, e_j$ are the embeddings of the extracted key elements for story $i$ and story $j$, respectively. $\cos(e_i, e_j)$ denotes their cosine similarity.
We set $K=5$ in this paper. 

\subsubsection{Story-Image Alignment Evaluator}
Inspired by prior text-image alignment research~\cite{vqascore2024}, we utilize a two-stage pipeline to evaluate the consistency between generated images and their stories.

In Key Point Extraction stage, we first utilize an LLM (GPT-4.1) as a structured extractor to identify key points across the seven predefined dimensions in Section~\ref{sec:corg}. 
Subsequently, in Alignment Scoring stage, we use an LVLM (GPT-4o) as the scorer to assesses the generated image against the structured key points extracted in the previous stage. The evaluation is multi-faceted: the scorer provides a fine-grained alignment score for each of the seven dimensions individually, as well as a holistic overall score summarizing the total alignment.

\subsection{Experiments}

\subsubsection{Experimental Settings}

\paragraph{Models}
We investigate different settings within the StorytellingPainter pipeline.
For the Storyteller model, we adopt five representative open-source LLMs with varying parameter sizes: Llama-3.2-1B-Instruct, Llama-3.2-3B-Instruct, Llama-3.1-8B-Instruct~\cite{grattafiori2024llama}, Qwen2.5-1.5B-Instruct and Qwen2.5-7B-Instruct~\cite{qwen2025qwen25technicalreport} and two of the state-of-the-art proprietary LLMs GPT-4o and GPT-4.1~\cite{Achiam2023GPT4TR}.
For the Painter model, we adopt two representative T2I models: DALL·E 3~\cite{betker2023improving} and GPT-image-1~\cite{gptimage1}.

\paragraph{Supplementary Evaluation Metrics}

Beyond the three aforementioned evaluators, we further assessed our method based on the following criteria: 

\begin{itemize}
    \item \textbf{Human Semantic Complexity Score}. 
    Although ISA model can automatically compute Semantic Score with relatively high alignment to human judgments, it still inevitably contains bias.
    Thus, to assess the semantic complexity of generated images more accurately, we also incorporate human evaluation. 
    \item \textbf{Number of Visual Clues}. 
    In the feature extraction stage, CoT VLISA (BERT) extracts seven dimensions of visual clues. 
    Thus, we calculate the number of visual clues based on the features as a reference. 
    \item \textbf{Number of Words}. 
    Since the generated stories are intended to be concise and suitable for image generation, rather than long narratives with temporal progression, we compute the number of words in each story as an indicator of instruction following. 
    
\end{itemize}

Details and effectiveness validation for our Evaluation Framework are provided in Appendix~\ref{sec:appendix_validations}. 

\begin{table*}[h!]
    \centering 
    \scriptsize 
    \setlength{\tabcolsep}{2.5pt} 
    \begin{tabular}{clcccccccccccc} 
    \hline 
    Mode & Model & Success Rate (\%) & \# Word & \# Visual Clue & Diversity &  Semantic Score  & Human  \\ 
    
    \hline
    & Llama-3.2-1B-Instruct  + DALL-E 3 & 100 & 384.33 & 7.00 & 0.382  & 17.2 & -- \\ 
    & Llama-3.2-1B-Instruct  + GPT-image-1$^\ast$ & 100 & 384.33 & 6.10 & 0.382  & 18.1 & -- \\ 
    & Llama-3.2-3B-Instruct  + DALL-E 3 & 100 &339.87 & 6.20 & 0.250 & 11.8 & -- \\ 
    & Llama-3.2-3B-Instruct  + GPT-image-1$^\ast$ & 100 & 339.87 & 6.50 &  0.250 & 10.0 & -- \\ 
    &  Llama-3.1-8B-Instruct + DALL-E 3 & 100 & 351.77 & 6.70 &  0.337 & 13.1 & -- \\ 
    &  Llama-3.1-8B-Instruct + GPT-image-1$^\ast$ & 96.7 & 351.77 & 6.38 &  0.337 & 13.0 & -- \\ 
    Naive & Qwen2.5-1.5B-Instruct+DALL·E 3 & 96.7  & 311.33 & 7.31 & 0.365  & 16.3 & -- \\ 
    & Qwen2.5-1.5B-Instruct+GPT-image-1$^\ast$ & 100  & 311.33 &  6.90 &  0.365 & 17.3 & -- \\ 
     & Qwen2.5-7B-Instruct+DALL·E 3 & 100  & 92.50 &  7.53 &  0.272 & 26.6 & -- \\ 
     & Qwen2.5-7B-Instruct+GPT-image-1$^\ast$ & 100  & 92.50 & 7.17 &  0.272 & 26.2 & -- \\ 
    & GPT-4o+DALL·E 3 & 100  & 144.07 & 8.37  & 0.312 & 34.1 & --  \\
    & GPT-4o+GPT-image-1$^\ast$ & 100  & 144.07 & 7.10 & 0.312 & 31.5 & -- \\  
    & GPT-4.1+DALL·E 3  & 100  & 111.37 & 8.37 & 0.262 & 39.1   & --  \\ 
    & GPT-4.1+GPT-image-1$^\ast$ & 100 & 111.37 & 8.53 & 0.262 & 40.7 & -- \\ 
    \hline
    & Llama-3.2-1B-Instruct+DALL-E 3 & 90.0 & 459.27 & 7.33 & \textbf{0.432} & 28.1 & --  \\
    & Llama-3.2-1B-Instruct  + GPT-image-1$^\ast$ & 96.7 & 459.27 & 7.17 &  \textbf{0.432} & 36.5 & 38.6 \\ 
    & Llama-3.2-3B-Instruct  + DALL-E 3 & 100 & 389.77 & 7.40 & 0.350  & 31.5 & -- \\ 
    & Llama-3.2-3B-Instruct  + GPT-image-1$^\ast$ & 100 & 389.77 & 7.80 &  0.350 & 36.3 & 35.6 \\ 
    &  Llama-3.1-8B-Instruct + DALL-E 3 & 100 & 359.27 & 7.37 & 0.374  & 25.2 & --  \\  
    &  Llama-3.1-8B-Instruct + GPT-image-1$^\ast$ & 100 & 359.27 & 7.53 & 0.374  & 34.2 &  32.1 \\ 
     & Qwen2.5-1.5B-Instruct+DALL·E 3 & 93.3  & 443.20 & 8.04 & \underline{0.407} & 29.1 & -- \\ 
     & Qwen2.5-1.5B-Instruct+GPT-image-1$^\ast$ & 100  & 443.20 &  8.13 & \underline{0.407}  & 31.1 & 29.9 \\ 
     CoR-Guided & Qwen2.5-1.5B-Instruct+GPT-image-1$^\dagger$ & 100  & 443.20 & 7.30 &  0.407 & 32.8 & 34.7 \\ 
    & Qwen2.5-7B-Instruct+DALL·E 3 & 100  & 165.83 &  7.83 &  0.381 & 34.7 & -- \\ 
      & Qwen2.5-7B-Instruct+GPT-image-1$^\ast$ & 100  & 165.83 & 7.87 & 0.381  & 37.2 & 37.8 \\ 
     & Qwen2.5-7B-Instruct+GPT-image-1$^\dagger$ & 100 & 165.83 & 8.07 & 0.381 & 38.9 & 44.6 \\ 
     & GPT-4o+DALL·E 3 & 100  & 179.10 & 8.17 & 0.393   & 49.3  &  -- \\ 
    & GPT-4o+GPT-image-1$^\ast$ & 100   & 179.10 & \textbf{9.00} & 0.393 & 52.4 &  \underline{57.6} \\ 
    & GPT-4o+GPT-image-1$^\dagger$ & 100   & 179.10 & 8.53 & 0.393  & 52.1 & \underline{57.6}   \\ 
    & GPT-4.1+DALL·E 3  &  100 & 183.13 & 8.83 & 0.367 & \underline{56.8} & --  \\ 
    & GPT-4.1+GPT-image-1$^\ast$ & 100  & 183.13 & 8.37 & 0.367 & 56.7  & 52.4  \\ 
    & GPT-4.1+GPT-image-1$^\dagger$ & 100  & 183.13 & \underline{8.93} & 0.367 & \textbf{61.0}  &  \textbf{62.9} \\ 
    \hline
     & CogBench in ISA test set & -- & 119.90 & 8.83 & 0.425 &  57.6  & 66.7  \\
     & CogBench in ISA test set$^{\uparrow30}$ & --  & 135.93 & 9.13 & 0.441  & 59.2  &   71.8  \\  
    \hline
    \end{tabular}
    \caption{
    \label{tab:sig_pipe_score}
    Performance of StorytellingPainter on Storytelling Image Generation. 
    Success Rate refers to the percentage of images successfully generated by the T2I model based on the story.
    Human refers to Human Semantic Score. 
    Numbers of Semantic Score are presented in \% with a full score of 100\%.
    ``$\ast$'' and ``$\dagger$'' denote ``medium'' and ``auto'' quality settings, respectively. 
    ``$\uparrow30$'' indicates the selection of the top-30 samples with the highest human evaluation scores.
    The highest scores are shown in bold, and the second-highest scores are underlined.
    }
\end{table*}

\paragraph{Implementation Details}

For each configuration, we generate 30 story-image pairs and report the averaged metrics as the final result. 
CogBench contains many Storytelling Images created by human illustrators, which we adopt as a reference for human capability. 
Considering that some CogBench images are included in the training data of the ISA model, we identified 48 overlapping images from the ISA test set to serve as our reference set. 
For CoR-Guided Mode, we split CogBench in a 3:2 ratio, with the two parts reserved for constructing training data and for testing, respectively.

\subsubsection{Results}

\paragraph{Comparison of prompting modes.}

Table~\ref{tab:sig_pipe_score} presents the performance across various configurations of the StorytellingPainter pipeline. 
By comparing the Semantic Score of images generated under two different prompting modes, we can observe that CoR-Guided Mode significantly improves the semantic complexity of generated stories and corresponding images. 
Specifically, according to the number of visual clues in Table~\ref{tab:sig_pipe_score} and Figure~\ref{fig:radar} (Appendix~\ref{sec:vc}), we can see that CoR-Guided Mode indeed improves the number of visual clues across multiple dimensions compared to the Naive Mode. 
In terms of diversity, the CoR-Guided Mode also achieves significantly higher Diversity Scores than the Naive Mode, which can be attributed to our Dyn-ICL strategy.

\paragraph{Performance of different Storyteller models.}
In Table~\ref{tab:sig_pipe_score}, by fixing the prompting mode and Painter model, we can evaluate the storytelling capabilities of different Storyteller models.

The combination of GPT-4o and GPT-4.1 as Storytellers with the GPT-Image-1 Painter yields the best performance. 
A comparison of their Semantic Scores against CogBench demonstrates their capability to generate plausible storytelling images. 
However, the score disparity between these models and the top 30 human-created images in CogBench indicates that their capabilities still fall significantly short of top-tier human performance.

The performance of smaller open-source models is significantly worse than GPT-4. 
The word count of stories generated by Qwen2.5-1.5B-Instruct, Llama-3.2-1B-Instruct, and Llama-3.2-3B-Instruct is substantially higher than that of other Storyteller models.
This is because they fail to understand the instructions and generate long stories with time sequences. 
Another piece of evidence is the fact that although Llama-3.2-1B-Instruct achieves higher Diversity and Semantic Scores, it is actually because it plagiarized many of the examples we provided for ICL. 
Other larger open-source models, while capable of understanding our task instructions and occasionally generating detailed stories, are still not very good at constructing narratives that are both logically coherent and semantically rich. 
This demonstrates the highly challenging nature of our task. 
Appendix~\ref{case:pipeline} presents story-image pairs generated by different Storytellers.

\paragraph{Influence of Painter model.}

\begin{figure}[h!]
    \centering
    \includegraphics[width=0.48\textwidth]{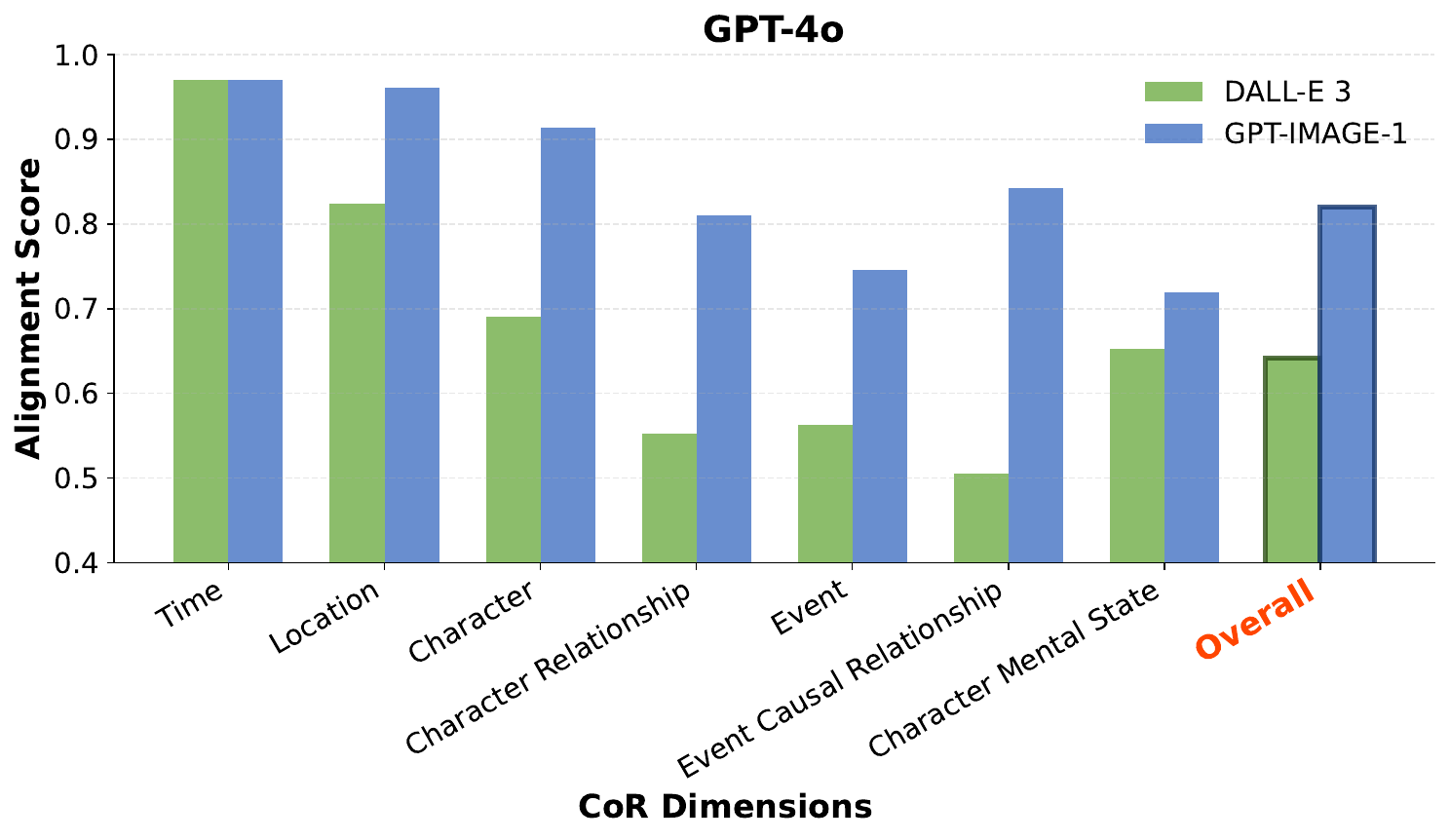}
    \caption{Story-image alignment evaluation of Painter models. The Storyteller model is GPT-4o.}
    \label{fig:alignment_eval}
\end{figure}

In terms of Painter models, as shown in Table~\ref{tab:sig_pipe_score}, for the same story, images generated by different Painter models yield different Semantic Scores. 
GPT-Image-1, a more recent T2I model, demonstrates better performance than DALL·E 3. 
When the ``quality'' hyperparameter of GPT-Image-1 is changed from ``medium'' to ``auto'', a significant improvement in performance can also be observed. 

This is attributed to the varying text-alignment capabilities of the Painter models. 
Higher-quality images generated by better Painter models can more accurately depict events, character emotions, other visual clues and their inter-connections in the image, so that ISA model or human can more easily understand the stories happening in the image. 
This conclusion can be validated by Figure~\ref{fig:alignment_eval}.
We can observe that GPT-image-1's Story-Image Alignment Score is significantly higher than that of DALL-E 3. 
The performance gap is particularly significant in [Character], [Character Relationship], [Event], and [Event Causal Relationship]. 
The significance of Painter is further corroborated by the case studies in Appendix~\ref{sec:case_painter}. 
This indicates our task imposes stringent requirements on T2I capabilities to deliver complex storytelling in one image.

\section{Mini-Storyteller}

In StorytellingPainter, the Storyteller acts as the brain to imagine stories. 
While proprietary LLMs excel in storytelling, they still fall short of expert human creators. 
Moreover, smaller open-source models struggle to match the creativity and reasoning of their proprietary counterparts.
Inspired by~\citet{eldan2023tinystoriessmalllanguagemodels}, as storytelling relies on language essentials like grammar and reasoning, Small Language Models (SLMs) have the potential to produce coherent narratives. 
Consequently, we explore fine-tuning strategies to enhance open-source models, yielding the lightweight Mini-Storytellers.

\subsection{Method}

As demonstrated in Section~\ref{sec:pipeline}, Qwen2.5-1.5B-Instruct and the Llama-3.2 series (1B and 3B) struggle to follow instructions for generating stories without a time sequence. 
Conversely, while larger open-source models adhere to instructions, their narratives remain unengaging. 
Additionally, our StorytellingPainter pipeline naturally functions as a data generator. 
Based on the observations above, we propose leveraging GPT-4.1 acting as the teacher Storyteller to generate high-quality training samples, thereby transferring its capabilities to smaller models via knowledge distillation. 
We investigated two common fine-tuning strategies: Supervised Fine-Tuning (SFT) and Direct Preference Optimization (DPO)~\cite{dpo}. 
Details are provided in Appendix~\ref{sec:loss}.

\paragraph{SFT}
The student model employs a standard Causal Language Modeling objective to maximize the likelihood of the target response. 
Therefore, we directly utilize the high-quality dataset of instruction-response pairs synthesized by GPT-4.1.

\paragraph{DPO}

To construct preference pairs for DPO, we designate student outputs as rejected and teacher outputs as chosen, thereby encouraging the student to align with the teacher's superior quality.
Additionally, for the larger Llama-3.1-8B and Qwen2.5-7B models, we explore utilizing outputs from their smaller counterparts as additional rejected samples. 
We term this strategy DPO-Mix. 
Specifically, the rejected set for Qwen2.5-7B includes both its own outputs and those from Qwen2.5-1.5B, while Llama-3.1-8B utilizes negatives from Llama-3.2-1B. 
For every preference pair, both chosen and rejected samples are generated from the same prompt to ensure direct comparability.

\begin{table*}[h!]
    \centering 
    \scriptsize 
    \begin{tabular}{lcccccc} 
    \hline 
    Model & Success Rate (\%) & \# Word & \# Visual Clue & Diversity & Semantic Score & Human  \\
    \hline
    Llama-3.2-1B-Instruct  & 96.7 & 459.27 & 7.33 & \textbf{0.432} &  36.5 & 38.6 \\ 
    Mini-Storyteller-LLaMA-3.2-1B-SFT  & 100 & 172.13 & \textbf{8.70} & 0.344 & \textbf{57.2} & 45.4 \\ 
    Mini-Storyteller-LLaMA-3.2-1B-DPO  & 96.7 & 79.97 & 7.10 & 0.381 & 47.8 & 37.2 \\ 
    \hdashline
    Llama-3.2-3B-Instruct & 100 & 389.77 & 7.80 & 0.350 &  36.3 & 35.6 \\ 
    Mini-Storyteller-LLaMA-3.2-3B-SFT   & 100 & 322.60 & 8.13 & 0.395 & 46.0 &  41.4 \\ 
    Mini-Storyteller-LLaMA-3.2-3B-DPO   & 100 & 289.13 & 7.27 & 0.399 & 41.5 & 39.3 \\ 
    \hdashline
    Llama-3.1-8B-Instruct  & 100 & 359.27 & 7.53 & 0.374  & 34.2 & 32.1 \\ 
    Mini-Storyteller-LLaMA-3.1-8B-SFT   & 100 & 233.70 & 7.97  &  0.389 &  49.4  &  \textbf{49.4} \\ 
    Mini-Storyteller-LLaMA-3.1-8B-DPO   & 100 & 182.83 & 7.70 & 0.390 & 49.2 & 43.3 \\ 
    Mini-Storyteller-LLaMA-3.1-8B-DPO-Mix   & 100 &  112.43 & 7.30 &  0.395 &  \underline{51.4}  &  \underline{47.2} \\ 
    \hdashline
    Qwen2.5-1.5B-Instruct & 100 & 443.20 & 8.13 & 0.407 & 31.1 & 29.9 \\ 
    Mini-Storyteller-Qwen2.5-1.5B-SFT  & 100 & 164.23 & 8.33 &  0.334 & 46.3 & 36.8 \\ 
    Mini-Storyteller-Qwen2.5-1.5B-DPO   & 100 & 201.17 & 7.83 & \underline{0.418} & 40.4 & 38.6 \\  
    \hdashline
    Qwen2.5-7B-Instruct  & 100 & 165.83 & 7.87 & 0.381 & 37.2 & 37.8 \\ 
    Mini-Storyteller-Qwen2.5-7B-SFT  & 100 & 157.77 & 8.23  &  0.342  & 45.2 & 42.9 \\ 
    Mini-Storyteller-Qwen2.5-7B-DPO   & 100 & 221.43 &  7.73 & 0.394 &  46.3 & 42.4 \\ 
    Mini-Storyteller-Qwen2.5-7B-DPO-Mix   & 100 & 153.63 & \underline{8.47} & 0.390 & 49.5 & 47.1 \\ 
    \hline
    \end{tabular}
    \caption{
    \label{tab:mini-storyteller}
    Performance of Mini-Storyteller models. The Painter model is GPT-image-1 with quality set to “medium”. The highest scores are shown in bold, and the second-highest scores are underlined. 
    }
\end{table*}

\subsection{Implementation Details}

We construct our dataset by operating the pipeline in CoR-Guided mode, utilizing prompts with examples sampled from the CogBench training split. 
We generated 2,000 samples using GPT-4.1 for SFT and DPO. 
Additionally, for the DPO-Mix training of Llama-3.1-8B-Instruct and Qwen2.5-7B-Instruct, we aggregated a 4,000-sample set by merging 2,000 outputs from their smaller models with 2,000 self-generated samples.
We train our models using LLaMA-Factory on Tesla V100 GPUs. 
Each model is fine-tuned using LoRA~\cite{hu2022lora} for one epoch with a total batch size of 16. 
We apply a cosine learning rate scheduler with a warmup ratio of 0.1. 
During the training process, we set aside 10\% of the training data for validation.

\subsection{Results}

Table~\ref{tab:mini-storyteller} reveals a significant increase in semantic complexity, demonstrating SFT and DPO effectively enhance semantically rich generation with StorytellingPainter data. 
Mini-Storyteller-LLaMA-3.1-8B series show better performance. 
Mini-Storyteller-LLaMA-3.1-8B-SFT achieves the highest Human Semantic Score, surpassing Llama-3.1-8B-Instruct by 15.2 automated and 17.3 human points.
Also, smaller Mini-Storyteller-LLaMA-3.2-1B-SFT achieves high Semantic Scores, validating the potential of small-scale models to generate fluent stories. 
Furthermore, word count for Llama-3.2-1B/3B-Instruct and Qwen2.5-1.5B-Instruct decreased significantly. 
This indicates they successfully followed instructions to generate concise stories, rather than long chronological narratives.

SFT holds a slight advantage over vanilla DPO in terms of semantic complexity. 
However, it is more likely to diminish output diversity, which can be observed by the Diversity Scores of Mini-Storyteller-LLaMA-3.2-1B-SFT and Mini-Storyteller-Qwen2.5-1.5B/7B-SFT.
For Llama-3.1-8B-Instruct and Qwen2.5-7B-Instruct, DPO-Mix shows better performance than vanilla DPO by introducing rejected samples from smaller-scale models. 
It can further improve the Semantic Score while preserving diversity.

\subsection{Case Analysis}

\begin{figure}[h]
    \centering
    \includegraphics[width=0.48\textwidth]{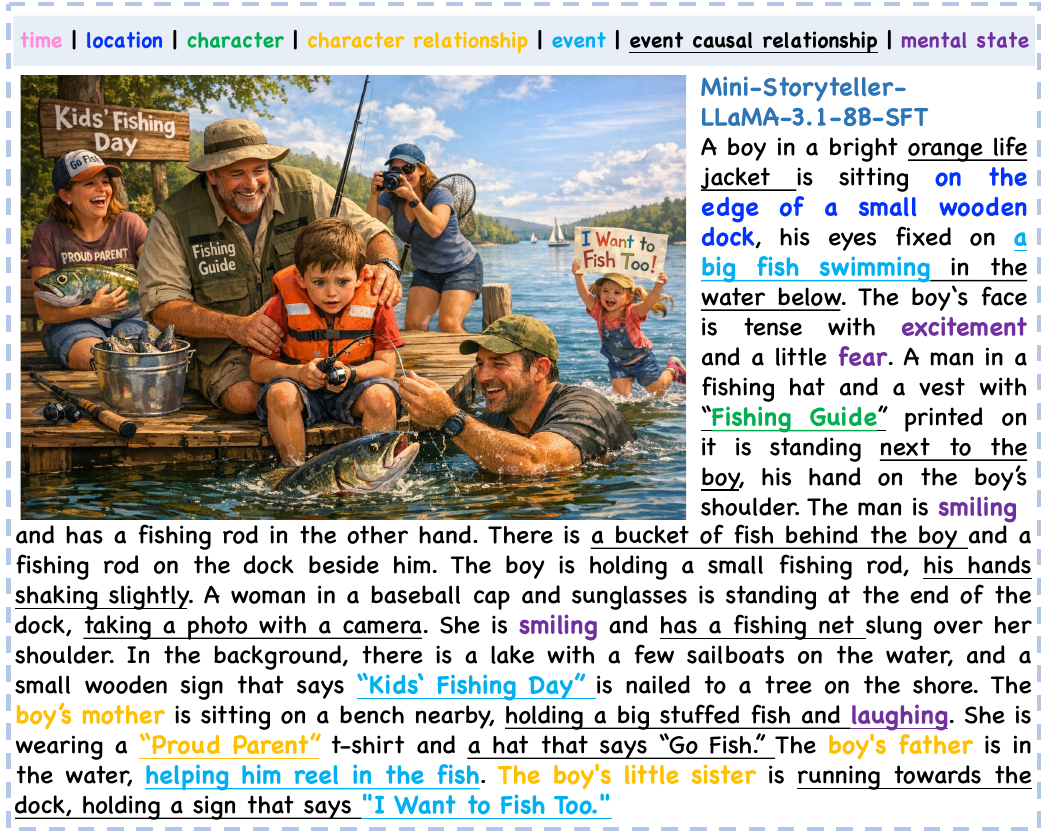}
    \caption{A story–image pair generated by StorytellingPainter with Mini-Storyteller-LLaMA-3.1-8B-SFT.
    }
    \label{fig:mini_best_case}
\end{figure}

Figure~\ref{fig:mini_best_case} illustrates a story generated by a Mini-Storyteller. 
The image contains rich visual clues forming CoRs across semantic dimensions like [Character Role], [Character Relationship], [Event], [Event Causal Relationship], and [Mental State].
``Kids' Fishing Day'' sets the scene as a boy hooks a big fish, aided by the guide and his father. While a woman photographs the moment and his mother laughs, the excitement makes the little sister eager to join in.
The bucket also implies a successful catch. 
More cases are in Appendix~\ref{sec:case_mini}.

\section{Related Work}

\paragraph{Story Generation}
Story Generation is a well-established research problem~\cite{roemmele2016writing, fan-etal-2018-hierarchical}. 
TinyStories~\cite{eldan2023tinystoriessmalllanguagemodels} shows SLMs are capable of generating fluent stories with basic reasoning.
STORYTELLER~\cite{li-etal-2025-storyteller} improves the engagement of stories by enhancing plot planning. 
HoLLMwood~\cite{chen-etal-2024-hollmwood} leverages LLMs for screenwriting. 
IBSEN~\cite{han-etal-2024-ibsen} generates drama scripts with multi-agent. 
Visual Storytelling task~\cite{huang2016visual} aims at generating a coherent paragraph-level story with the image stream as input. 
StoryLLaVA~\cite{yang-etal-2025-storyllava} generates narratives from sequential images by optimizing story data and employing a multi-phase training strategy. 
Unlike previous studies, our work on storytelling places greater emphasis on imagining a semantically rich and logically coherent story that takes place in a single moment and can be captured in a single image.

\paragraph{Story Visualization}
Story Visualization~\cite{8953914} generates image sequences from a sequence of story captions. 
Numerous studies~\cite{maharana2021improving, maharana2021integrating, maharana2022storydall} have investigated various methods. 
StoryImager~\cite{tao2024storyimager} proposes a unified, bidirectional framework to achieve coherent story visualization. 
StoryGPT-V~\cite{Shen_2025_CVPR} utilizes latent diffusion and LLM to generate images with consistent characters. 
SEED-Story~\cite{yang2024seedstory} leverages LVLMs to produce extended multimodal stories conditioned on visual and textual inputs. 
\citet{roemmele2025llmsscenesenablingnarrative} generate story illustrations by prompting LLMs to produce scene descriptions that guide T2I models. 
Their goal is to create auxiliary illustrations for existing stories, rather than standalone, semantically rich images.
Unlike previous tasks, Storytelling Image Generation targets a single, semantically dense image rather than a sequence of simpler images, and generates the story itself instead of taking it as input.

\section{Conclusion}

In this paper, we propose a novel Storytelling Image Generation task to generate semantically rich images conveying coherent stories. 
To this end, we present StorytellingPainter, a two-stage pipeline combining LLM creativity and T2I synthesis, along with a specialized evaluation framework. 
To narrow the storytelling gap with proprietary models, we leverage this pipeline to generate data and distill capabilities into small open-source LLMs, resulting in Mini-Storytellers. 
Storytelling Image Generation holds great potential for cognitive assessment, illustration, and enhancing multimodal models, yet remains challenging as it imposes stricter requirements on logical coherence and creativity from LLMs, as well as precise alignment from T2I models.

\section*{Limitations}

We present a preliminary exploration of Storytelling Image Generation in this paper. 
However, several challenges within this task remain to be addressed. 
Regarding semantic complexity, the logical connections in images generated by existing models are still insufficient.
Some model-generated stories lack logical soundness. 
Furthermore, concerning creative capability, the output diversity remains to be improved, as current models sometimes generate stories with repetitive themes. 
Addressing these challenges warrants further exploration in future work.

\section*{Ethical Considerations}

All data used in this paper were acquired through official channels in compliance with usage agreements and are used for research purposes only. 
All human evaluators were compensated fairly according to local standards. 
    %





\bibliography{custom}

@inproceedings{song-etal-2025-cognitive,
    title = "A Cognitive Evaluation Benchmark of Image Reasoning and Description for Large Vision-Language Models",
    author = "Song, Xiujie  and
      Wu, Mengyue  and
      Zhu, Kenny Q.  and
      Zhang, Chunhao  and
      Chen, Yanyi",
    editor = "Chiruzzo, Luis  and
      Ritter, Alan  and
      Wang, Lu",
    booktitle = "Proceedings of the 2025 Conference of the Nations of the Americas Chapter of the Association for Computational Linguistics: Human Language Technologies (Volume 1: Long Papers)",
    month = apr,
    year = "2025",
    address = "Albuquerque, New Mexico",
    publisher = "Association for Computational Linguistics",
    url = "https://aclanthology.org/2025.naacl-long.324/",
    pages = "6392--6409",
    ISBN = "979-8-89176-189-6"
}

@article{Song_Pang_Tang_Wu_Zhu_2025, 
    title={Is Your Image a Good Storyteller?}, 
    volume={39}, 
    url={https://ojs.aaai.org/index.php/AAAI/article/view/34702}, 
    DOI={10.1609/aaai.v39i24.34702}, 
    abstractNote={Quantifying image complexity at the entity level is straightforward, but the assessment of semantic complexity has been largely overlooked. In fact, there are differences in semantic complexity across images. Images with richer semantics can tell vivid and engaging stories and offer a wide range of application scenarios. For example, the Cookie Theft picture is such a kind of image and is widely used to assess human language and cognitive abilities due to its higher semantic complexity. Additionally, semantically rich images can benefit the development of vision models, as images with limited semantics are becoming less challenging for them. However, such images are scarce, highlighting the need for a greater number of them. For instance, there is a need for more images like Cookie Theft to cater to people from different cultural backgrounds and eras. Assessing semantic complexity requires human experts and empirical evidence. Automatic evaluation of how semantically rich an image will be the first step of mining or generating more images with rich semantics, and benefit human cognitive assessment, Artificial Intelligence, and various other applications. In response, we propose the Image Semantic Assessment (ISA) task to address this problem. We introduce the first ISA dataset and a novel method that leverages language to solve this vision problem. Experiments on our dataset demonstrate the effectiveness of our approach.}, 
    number={24}, 
    journal={Proceedings of the AAAI Conference on Artificial Intelligence}, 
    author={Song, Xiujie and Pang, Xiaoyi and Tang, Haifeng and Wu, Mengyue and Zhu, Kenny Q.}, year={2025}, month={Apr.}, pages={25165-25173} 
}

@misc{eldan2023tinystoriessmalllanguagemodels,
      title={TinyStories: How Small Can Language Models Be and Still Speak Coherent English?}, 
      author={Ronen Eldan and Yuanzhi Li},
      year={2023},
      eprint={2305.07759},
      archivePrefix={arXiv},
      primaryClass={cs.CL},
      url={https://arxiv.org/abs/2305.07759}, 
}

@INPROCEEDINGS{8953914,
  author={Li, Yitong and Gan, Zhe and Shen, Yelong and Liu, Jingjing and Cheng, Yu and Wu, Yuexin and Carin, Lawrence and Carlson, David and Gao, Jianfeng},
  booktitle={2019 IEEE/CVF Conference on Computer Vision and Pattern Recognition (CVPR)}, 
  title={StoryGAN: A Sequential Conditional GAN for Story Visualization}, 
  year={2019},
  volume={},
  number={},
  pages={6322-6331},
  doi={10.1109/CVPR.2019.00649}}

@article{betker2023improving,
  title={Improving image generation with better captions},
  author={Betker, James and Goh, Gabriel and Jing, Li and Brooks, Tim and Wang, Jianfeng and Li, Linjie and Ouyang, Long and Zhuang, Juntang and Lee, Joyce and Guo, Yufei and others},
  journal={Computer Science. https://cdn. openai. com/papers/dall-e-3. pdf},
  volume={2},
  number={3},
  pages={8},
  year={2023}
}

@BOOK{goodglass2001-ej,
  title     = "{BDAE}: The Boston Diagnostic Aphasia Examination",
  author    = "Goodglass, Harold and Kaplan, Edith and Weintraub, Sandra",
  publisher = "Lippincott Williams \& Wilkins",
  year      =  2001,
  address   = "Philadelphia, PA"
}

@article{berube2019stealing,
  title={Stealing cookies in the twenty-first century: Measures of spoken narrative in healthy versus speakers with aphasia},
  author={Berube, Shauna and Nonnemacher, Jodi and Demsky, Cornelia and Glenn, Shenly and Saxena, Sadhvi and Wright, Amy and Tippett, Donna C and Hillis, Argye E},
  journal={American journal of speech-language pathology},
  volume={28},
  number={1S},
  pages={321--329},
  year={2019},
  publisher={ASHA}
}

@article{rethinkingct,
author = {Amy Steinberg  and Patrick D. Lyden  and Arielle P. Davis },
title = {Bias in Stroke Evaluation: Rethinking the Cookie Theft Picture},
journal = {Stroke},
volume = {53},
number = {6},
pages = {2123-2125},
year = {2022},
doi = {10.1161/STROKEAHA.121.038515}}

@article{HUSSEIN2015152,
title = {Arabic cross cultural adaptation and validation of the National Institutes of Health Stroke Scale},
journal = {Journal of the Neurological Sciences},
volume = {357},
number = {1},
pages = {152-156},
year = {2015},
issn = {0022-510X},
doi = {https://doi.org/10.1016/j.jns.2015.07.022},
url = {https://www.sciencedirect.com/science/article/pii/S0022510X15004438},
author = {Haitham M. Hussein and Amr {Abdel Moneim} and Tamer Emara and Yousry A. Abd-elhamid and Haitham H. Salem and Foad Abd-Allah and Mohammad A. Farrag and M. Amir Tork and Ali S. Shalash and Khaled H. {Ezz el dein} and Gamaleldin Osman and Shady S. Georgy and Peter G. Ghali and Patrick D. Lyden and Ramez R. Moustafa}
}

@article{DOMINGUEZ2006476,
title = {Spanish Cross-Cultural Adaptation and Validation of the National Institutes of Health Stroke Scale},
journal = {Mayo Clinic Proceedings},
volume = {81},
number = {4},
pages = {476-480},
year = {2006},
issn = {0025-6196},
doi = {https://doi.org/10.4065/81.4.476},
url = {https://www.sciencedirect.com/science/article/pii/S0025619611618958},
author = {Raúl Domínguez and José F. Vila and Federico Augustovski and Vilma Irazola and Pablo R. Castillo and Roberto Rotta Escalante and Thomas G. Brott and James F. Meschia}
}

@article{Oh2012ValidityAR,
  title={Validity and Reliability of a Korean Version of the National Institutes of Health Stroke Scale},
  author={Mi Sun Oh and Kyung Ho Yu and Ju-Hun Lee and San Jung and I S Ko and Joon Hyun Shin and Soo-Jin Cho and Hui-Chul Choi and Hyang Hee Kim and Byung‐Chul Lee},
  journal={Journal of Clinical Neurology (Seoul, Korea)},
  year={2012},
  volume={8},
  pages={177 - 183},
  url={https://api.semanticscholar.org/CorpusID:12210538}
}

@misc{qwen2025qwen25technicalreport,
      title={Qwen2.5 Technical Report}, 
      author={Qwen and : and An Yang and Baosong Yang and Beichen Zhang and Binyuan Hui and Bo Zheng and Bowen Yu and Chengyuan Li and Dayiheng Liu and Fei Huang and Haoran Wei and Huan Lin and Jian Yang and Jianhong Tu and Jianwei Zhang and Jianxin Yang and Jiaxi Yang and Jingren Zhou and Junyang Lin and Kai Dang and Keming Lu and Keqin Bao and Kexin Yang and Le Yu and Mei Li and Mingfeng Xue and Pei Zhang and Qin Zhu and Rui Men and Runji Lin and Tianhao Li and Tianyi Tang and Tingyu Xia and Xingzhang Ren and Xuancheng Ren and Yang Fan and Yang Su and Yichang Zhang and Yu Wan and Yuqiong Liu and Zeyu Cui and Zhenru Zhang and Zihan Qiu},
      year={2025},
      eprint={2412.15115},
      archivePrefix={arXiv},
      primaryClass={cs.CL},
      url={https://arxiv.org/abs/2412.15115}, 
}

@inproceedings{Achiam2023GPT4TR,
  title={GPT-4 Technical Report},
  author={OpenAI},
  year={2023},
  url={https://api.semanticscholar.org/CorpusID:257532815}
}

@article{yang2024seedstory,
      title={SEED-Story: Multimodal Long Story Generation with Large Language Model}, 
      author={Shuai Yang and Yuying Ge and Yang Li and Yukang Chen and Yixiao Ge and Ying Shan and Yingcong Chen},
      year={2024},
      journal={arXiv preprint arXiv:2407.08683},
      url={https://arxiv.org/abs/2407.08683}, 
}

@inproceedings{fan-etal-2018-hierarchical,
    title = "Hierarchical Neural Story Generation",
    author = "Fan, Angela  and
      Lewis, Mike  and
      Dauphin, Yann",
    editor = "Gurevych, Iryna  and
      Miyao, Yusuke",
    booktitle = "Proceedings of the 56th Annual Meeting of the Association for Computational Linguistics (Volume 1: Long Papers)",
    month = jul,
    year = "2018",
    address = "Melbourne, Australia",
    publisher = "Association for Computational Linguistics",
    url = "https://aclanthology.org/P18-1082/",
    doi = "10.18653/v1/P18-1082",
    pages = "889--898"
}

@InProceedings{Shen_2025_CVPR,
    author    = {Shen, Xiaoqian and Elhoseiny, Mohamed},
    title     = {StoryGPT-V: Large Language Models as Consistent Story Visualizers},
    booktitle = {Proceedings of the Computer Vision and Pattern Recognition Conference (CVPR)},
    month     = {June},
    year      = {2025},
    pages     = {13273-13283}
}

@article{maharana2021improving,
  title={Improving generation and evaluation of visual stories via semantic consistency},
  author={Maharana, Adyasha and Hannan, Darryl and Bansal, Mohit},
  journal={arXiv preprint arXiv:2105.10026},
  year={2021}
}

@inproceedings{maharana2021integrating,
  title={Integrating Visuospatial, Linguistic and Commonsense Structure into Story Visualization},
  author={Maharana, Adyasha and Bansal, Mohit},
  booktitle={EMNLP},
  year={2021}
}

@inproceedings{maharana2022storydall,
  title={Storydall-e: Adapting pretrained text-to-image transformers for story continuation},
  author={Maharana, Adyasha and Hannan, Darryl and Bansal, Mohit},
  booktitle={European Conference on Computer Vision},
  pages={70--87},
  year={2022},
  organization={Springer}
}

@inproceedings{yang-etal-2025-storyllava,
    title = "{S}tory{LL}a{VA}: Enhancing Visual Storytelling with Multi-Modal Large Language Models",
    author = "Yang, Li  and
      Xiao, Zhiding  and
      Huang, Wenxin  and
      Zhong, Xian",
    editor = "Rambow, Owen  and
      Wanner, Leo  and
      Apidianaki, Marianna  and
      Al-Khalifa, Hend  and
      Eugenio, Barbara Di  and
      Schockaert, Steven",
    booktitle = "Proceedings of the 31st International Conference on Computational Linguistics",
    month = jan,
    year = "2025",
    address = "Abu Dhabi, UAE",
    publisher = "Association for Computational Linguistics",
    url = "https://aclanthology.org/2025.coling-main.266/",
    pages = "3936--3951"
}

@inproceedings{huang2016visual,
  title={Visual storytelling},
  author={Huang, Ting-Hao and Ferraro, Francis and Mostafazadeh, Nasrin and Misra, Ishan and Agrawal, Aishwarya and Devlin, Jacob and Girshick, Ross and He, Xiaodong and Kohli, Pushmeet and Batra, Dhruv and others},
  booktitle={Proceedings of the 2016 conference of the North American chapter of the association for computational linguistics: Human language technologies},
  pages={1233--1239},
  year={2016}
}

@inproceedings{chen-etal-2024-hollmwood,
    title = "{H}o{LLM}wood: Unleashing the Creativity of Large Language Models in Screenwriting via Role Playing",
    author = "Chen, Jing  and
      Zhu, Xinyu  and
      Yang, Cheng  and
      Shi, Chufan  and
      Xi, Yadong  and
      Zhang, Yuxiang  and
      Wang, Junjie  and
      Pu, Jiashu  and
      Feng, Tian  and
      Yang, Yujiu  and
      Zhang, Rongsheng",
    editor = "Al-Onaizan, Yaser  and
      Bansal, Mohit  and
      Chen, Yun-Nung",
    booktitle = "Findings of the Association for Computational Linguistics: EMNLP 2024",
    month = nov,
    year = "2024",
    address = "Miami, Florida, USA",
    publisher = "Association for Computational Linguistics",
    url = "https://aclanthology.org/2024.findings-emnlp.474/",
    doi = "10.18653/v1/2024.findings-emnlp.474",
    pages = "8075--8121"
}

@inproceedings{han-etal-2024-ibsen,
    title = "{IBSEN}: Director-Actor Agent Collaboration for Controllable and Interactive Drama Script Generation",
    author = "Han, Senyu  and
      Chen, Lu  and
      Lin, Li-Min  and
      Xu, Zhengshan  and
      Yu, Kai",
    booktitle = "Proceedings of the 62nd Annual Meeting of the Association for Computational Linguistics (Volume 1: Long Papers)",
    month = aug,
    year = "2024",
    address = "Bangkok, Thailand",
    publisher = "Association for Computational Linguistics",
    url = "https://aclanthology.org/2024.acl-long.88",
    pages = "1607--1619",
}

@misc{gptimage1,
  author       = {OpenAI},
  title        = {GPT Image 1: State-of-the-art image generation model},
  howpublished = {\url{https://platform.openai.com/docs/models/gpt-image-1}},
  year         = {2025}
}

@inproceedings{dpo,
author = {Rafailov, Rafael and Sharma, Archit and Mitchell, Eric and Ermon, Stefano and Manning, Christopher D. and Finn, Chelsea},
title = {Direct preference optimization: your language model is secretly a reward model},
year = {2023},
publisher = {Curran Associates Inc.},
address = {Red Hook, NY, USA},
booktitle = {Proceedings of the 37th International Conference on Neural Information Processing Systems},
articleno = {2338},
numpages = {14},
location = {New Orleans, LA, USA},
series = {NIPS '23}
}

@article{hu2022lora,
  title={Lora: Low-rank adaptation of large language models.},
  author={Hu, Edward J and Shen, Yelong and Wallis, Phillip and Allen-Zhu, Zeyuan and Li, Yuanzhi and Wang, Shean and Wang, Lu and Chen, Weizhu and others},
  journal={ICLR},
  volume={1},
  number={2},
  pages={3},
  year={2022}
}

@inproceedings{li-etal-2025-storyteller,
    title = "{STORYTELLER}: An Enhanced Plot-Planning Framework for Coherent and Cohesive Story Generation",
    author = "Li, Jiaming  and
      Chen, Yukun  and
      Liu, Ziqiang  and
      Tan, Minghuan  and
      Zhang, Lei  and
      Li, Yunshui  and
      Luo, Run  and
      Chen, Longze  and
      Luo, Jing  and
      Argha, Ahmadreza  and
      Alinejad-Rokny, Hamid  and
      Zhou, Wei  and
      Yang, Min",
    editor = "Che, Wanxiang  and
      Nabende, Joyce  and
      Shutova, Ekaterina  and
      Pilehvar, Mohammad Taher",
    booktitle = "Findings of the Association for Computational Linguistics: ACL 2025",
    month = jul,
    year = "2025",
    address = "Vienna, Austria",
    publisher = "Association for Computational Linguistics",
    url = "https://aclanthology.org/2025.findings-acl.1071/",
    pages = "20818--20846",
    ISBN = "979-8-89176-256-5"
}

@inproceedings{tao2024storyimager,
  title={Storyimager: A unified and efficient framework for coherent story visualization and completion},
  author={Tao, Ming and Bao, Bing-Kun and Tang, Hao and Wang, Yaowei and Xu, Changsheng},
  booktitle={European Conference on Computer Vision},
  pages={479--495},
  year={2024},
  organization={Springer}
}

@article{nicholas1993system,
  title={A system for quantifying the informativeness and efficiency of the connected speech of adults with aphasia},
  author={Nicholas, Linda E and Brookshire, Robert H},
  journal={Journal of Speech, Language, and Hearing Research},
  volume={36},
  number={2},
  pages={338--350},
  year={1993},
  publisher={American Speech-Language-Hearing Association}
}

@article{describe-ctp,
author = {Cummings, Louise},
year = {2019},
month = {03},
pages = {151-174},
title = {Describing the Cookie Theft picture: Sources of breakdown in Alzheimer's dementia},
volume = {10},
journal = {Pragmatics and Society},
doi = {10.1075/ps.17011.cum}
}

@misc{roemmele2025llmsscenesenablingnarrative,
      title={LLMs Behind the Scenes: Enabling Narrative Scene Illustration}, 
      author={Melissa Roemmele and John Joon Young Chung and Taewook Kim and Yuqian Sun and Alex Calderwood and Max Kreminski},
      year={2025},
      eprint={2509.22940},
      archivePrefix={arXiv},
      primaryClass={cs.CL},
      url={https://arxiv.org/abs/2509.22940}, 
}

@inproceedings{yang-etal-2025-measuring,
    title = "Measuring Data Diversity for Instruction Tuning: A Systematic Analysis and A Reliable Metric",
    author = "Yang, Yuming  and
      Nan, Yang  and
      Ye, Junjie  and
      Dou, Shihan  and
      Wang, Xiao  and
      Li, Shuo  and
      Lv, Huijie  and
      Gui, Tao  and
      Zhang, Qi  and
      Huang, Xuanjing",
    editor = "Che, Wanxiang  and
      Nabende, Joyce  and
      Shutova, Ekaterina  and
      Pilehvar, Mohammad Taher",
    booktitle = "Proceedings of the 63rd Annual Meeting of the Association for Computational Linguistics (Volume 1: Long Papers)",
    month = jul,
    year = "2025",
    address = "Vienna, Austria",
    publisher = "Association for Computational Linguistics",
    url = "https://aclanthology.org/2025.acl-long.908/",
    doi = "10.18653/v1/2025.acl-long.908",
    pages = "18530--18549",
    ISBN = "979-8-89176-251-0"
}

@inproceedings{stasaski-hearst-2022-semantic,
    title = "Semantic Diversity in Dialogue with Natural Language Inference",
    author = "Stasaski, Katherine  and
      Hearst, Marti",
    editor = "Carpuat, Marine  and
      de Marneffe, Marie-Catherine  and
      Meza Ruiz, Ivan Vladimir",
    booktitle = "Proceedings of the 2022 Conference of the North American Chapter of the Association for Computational Linguistics: Human Language Technologies",
    month = jul,
    year = "2022",
    address = "Seattle, United States",
    publisher = "Association for Computational Linguistics",
    url = "https://aclanthology.org/2022.naacl-main.6/",
    doi = "10.18653/v1/2022.naacl-main.6",
    pages = "85--98"
}

@inproceedings{stasaski-etal-2020-diverse,
    title = "More Diverse Dialogue Datasets via Diversity-Informed Data Collection",
    author = "Stasaski, Katherine  and
      Yang, Grace Hui  and
      Hearst, Marti A.",
    editor = "Jurafsky, Dan  and
      Chai, Joyce  and
      Schluter, Natalie  and
      Tetreault, Joel",
    booktitle = "Proceedings of the 58th Annual Meeting of the Association for Computational Linguistics",
    month = jul,
    year = "2020",
    address = "Online",
    publisher = "Association for Computational Linguistics",
    url = "https://aclanthology.org/2020.acl-main.446/",
    doi = "10.18653/v1/2020.acl-main.446",
    pages = "4958--4968"
}

@inproceedings{vqascore2024,
author = {Lin, Zhiqiu and Pathak, Deepak and Li, Baiqi and Li, Jiayao and Xia, Xide and Neubig, Graham and Zhang, Pengchuan and Ramanan, Deva},
title = {Evaluating Text-to-Visual Generation with Image-to-Text Generation},
year = {2024},
isbn = {978-3-031-72672-9},
publisher = {Springer-Verlag},
address = {Berlin, Heidelberg},
url = {https://doi.org/10.1007/978-3-031-72673-6_20},
doi = {10.1007/978-3-031-72673-6_20},
booktitle = {Computer Vision – ECCV 2024: 18th European Conference, Milan, Italy, September 29–October 4, 2024, Proceedings, Part IX},
pages = {366–384},
numpages = {19},
location = {Milan, Italy}
}

@article{grattafiori2024llama,
  title={The llama 3 herd of models},
  author={Grattafiori, Aaron and Dubey, Abhimanyu and Jauhri, Abhinav and Pandey, Abhinav and Kadian, Abhishek and Al-Dahle, Ahmad and Letman, Aiesha and Mathur, Akhil and Schelten, Alan and Vaughan, Alex and others},
  journal={arXiv preprint arXiv:2407.21783},
  year={2024}
}

@inproceedings{devlin-etal-2019-bert,
    title = "{BERT}: Pre-training of Deep Bidirectional Transformers for Language Understanding",
    author = "Devlin, Jacob  and
      Chang, Ming-Wei  and
      Lee, Kenton  and
      Toutanova, Kristina",
    editor = "Burstein, Jill  and
      Doran, Christy  and
      Solorio, Thamar",
    booktitle = "Proceedings of the 2019 Conference of the North {A}merican Chapter of the Association for Computational Linguistics: Human Language Technologies, Volume 1 (Long and Short Papers)",
    month = jun,
    year = "2019",
    address = "Minneapolis, Minnesota",
    publisher = "Association for Computational Linguistics",
    url = "https://aclanthology.org/N19-1423/",
    doi = "10.18653/v1/N19-1423",
    pages = "4171--4186"
}

@inproceedings{roemmele2016writing,
  title={Writing stories with help from recurrent neural networks},
  author={Roemmele, Melissa},
  booktitle={Proceedings of the AAAI Conference on Artificial Intelligence},
  volume={30},
  number={1},
  year={2016}
}

@book{claridge2001norman,
  title={Norman Rockwell},
  author={Claridge, Laura},
  year={2001},
  publisher={Random House}
}

\appendix

\clearpage

\section{Distribution of Visual Clues of StorytellingPainters}
\label{sec:vc}

\begin{figure}[h!]
    \centering
    \includegraphics[width=0.49\textwidth]{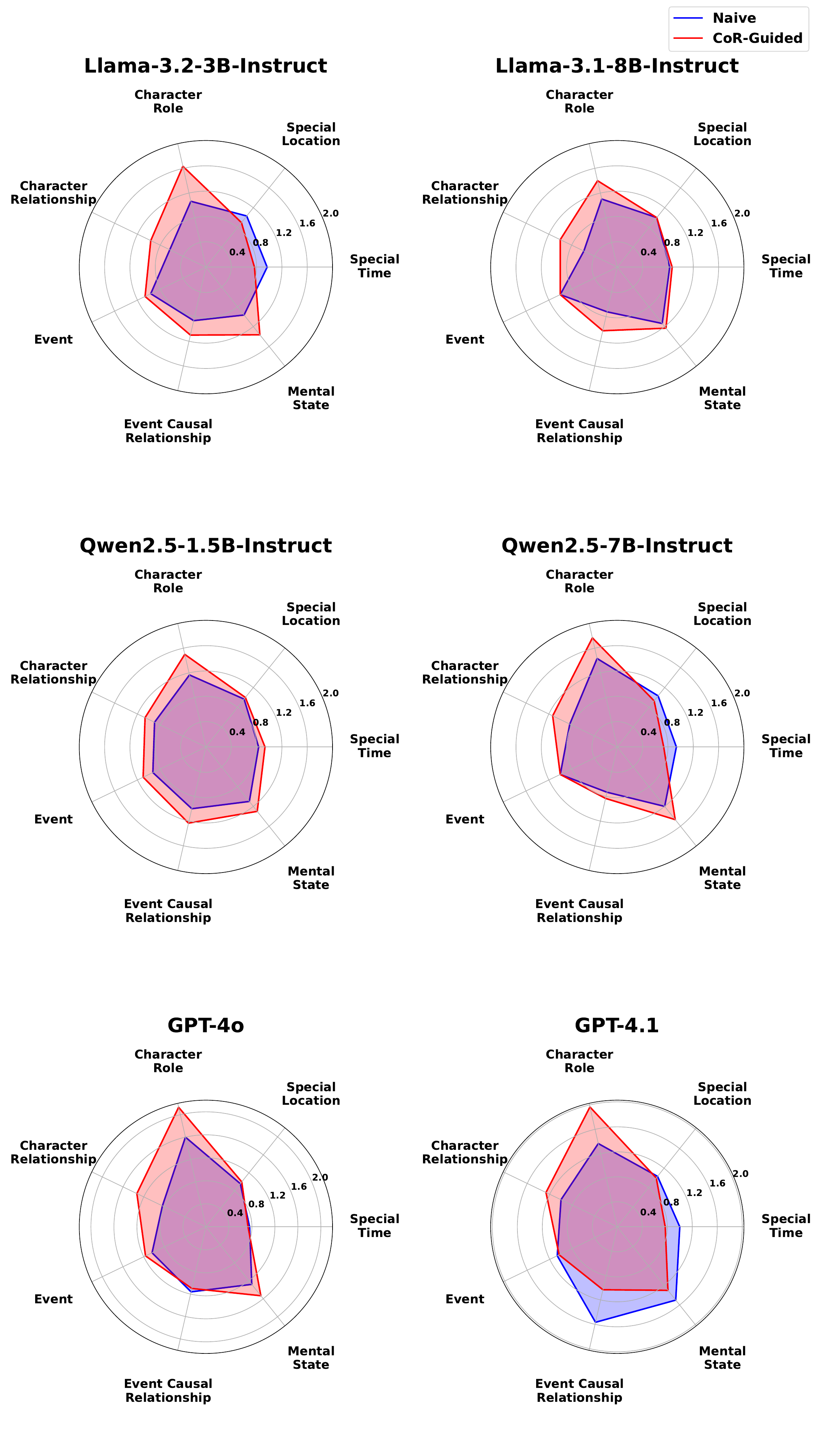}
    \caption{
    Distribution of visual clues across the seven dimensions in images generated by the StorytellingPainter pipeline with different Storyteller models. The Painter model is fixed as GPT-Image-1.
    }
    \label{fig:radar}
\end{figure}

Figure~\ref{fig:radar} illustrates the distribution of visual clues across various dimensions for images generated by different Storyteller models under Naive and CoR-Guided modes.
In general, it demonstrates that, compared to Naive Mode, CoR-Guided Mode effectively increases the number of visual clues in generated images across various dimensions. 

\section{Case Analysis of Storyteller Models}
\label{case:pipeline}

\begin{figure*}[h!]
    \centering
    \includegraphics[width=0.98\textwidth]{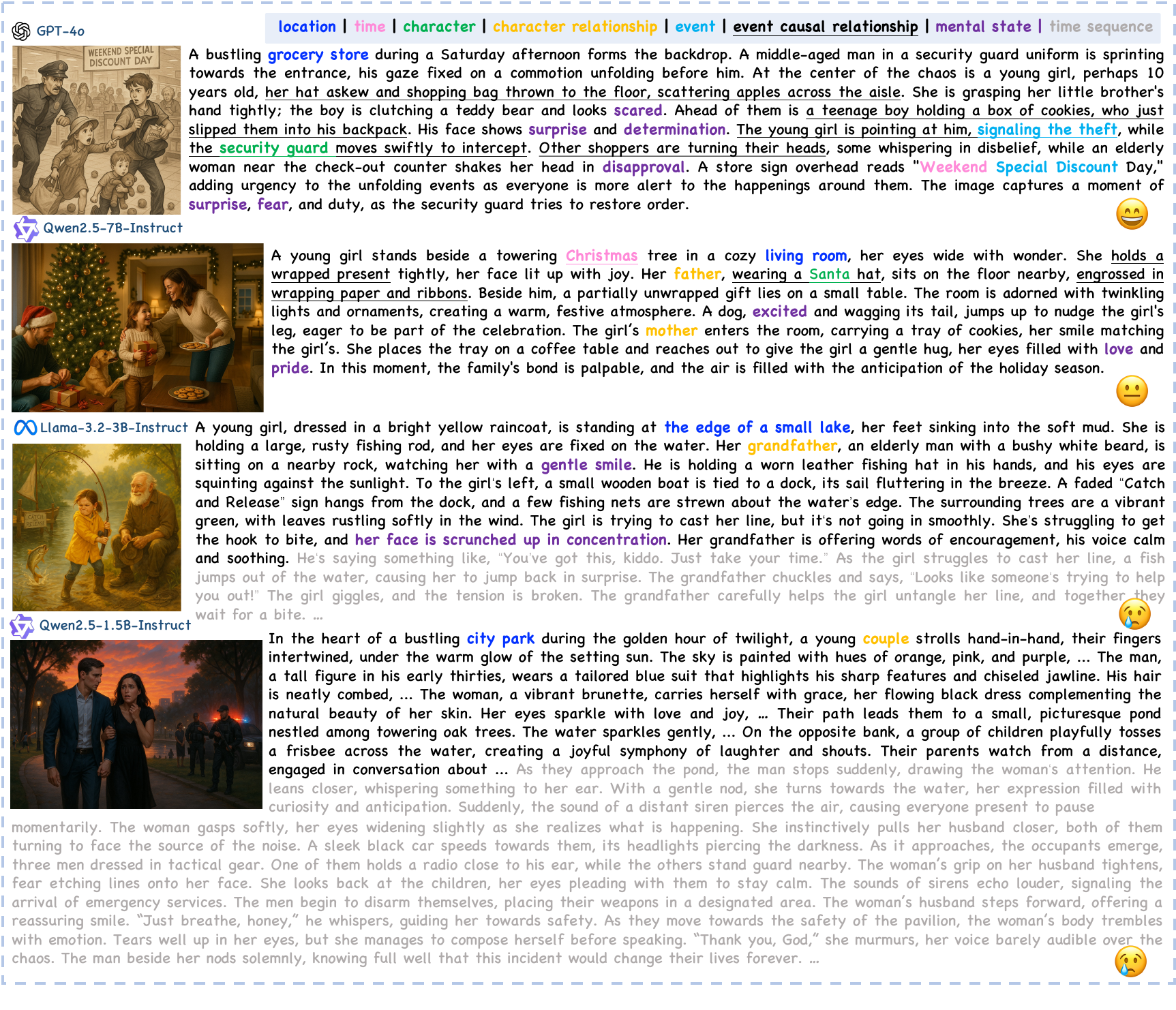}
    \caption{Stories and corresponding images generated by the StorytellingPainter pipeline. 
    }
    \label{fig:sig_case}
\end{figure*}

Figure~\ref{fig:sig_case} shows the stories and corresponding images generated by the StorytellingPainter pipeline. 
According to the cases, we can see that GPT-4o is capable of generating logically coherent and semantically rich stories and subsequently producing corresponding images. 
For example, the story generated by GPT-4o takes place in a grocery store during a weekend discount day, where a police officer apprehends a thief attempting to steal items.
The generated image contains relatively rich visual clues and CoRs. 
Again, as shown in the comparison between the image generated by GPT-4o and the one in Figure~\ref{fig:examples} (b), we observe that while these models can produce some qualified Storytelling Images, they still struggle to generate semantically richer images comparable to those created by outstanding human illustrators.

However, for open-source Storyteller models Qwen2.5-7B-Instruct, Llama-3.2-3B-Instruct and Qwen2.5-1.5B-Instruct, the stories they generate are noticeably less engaging and coherent in terms of both interest and logical consistency.
Though Qwen2.5-7B-Instruct can generate logically coherent stories, the semantic complexity is obviously lower. 
For the image generated by Qwen2.5-7B-Instruct, it merely describes a scene of a family celebrating Christmas, without presenting rich visual clues or causal relationships between events. 
Llama-3.2-3B-Instruct and Qwen2.5-1.5B-Instruct exhibit lower performance. 
The stories generated by them are usually primarily a stacking of multiple events. 
In the two cases, the models failed to follow the instructions properly by generating a long story with a temporal sequence and even dialogue, rather than a story that could be depicted within a single image. 
Overall, these cases highlight the critical role of Storytelling models in generating Storytelling Images, while also revealing their limitations.

\section{Case Analysis of Painter Models}
\label{sec:case_painter}
\begin{figure*}[h!]
    \centering
    \includegraphics[width=0.98\textwidth]{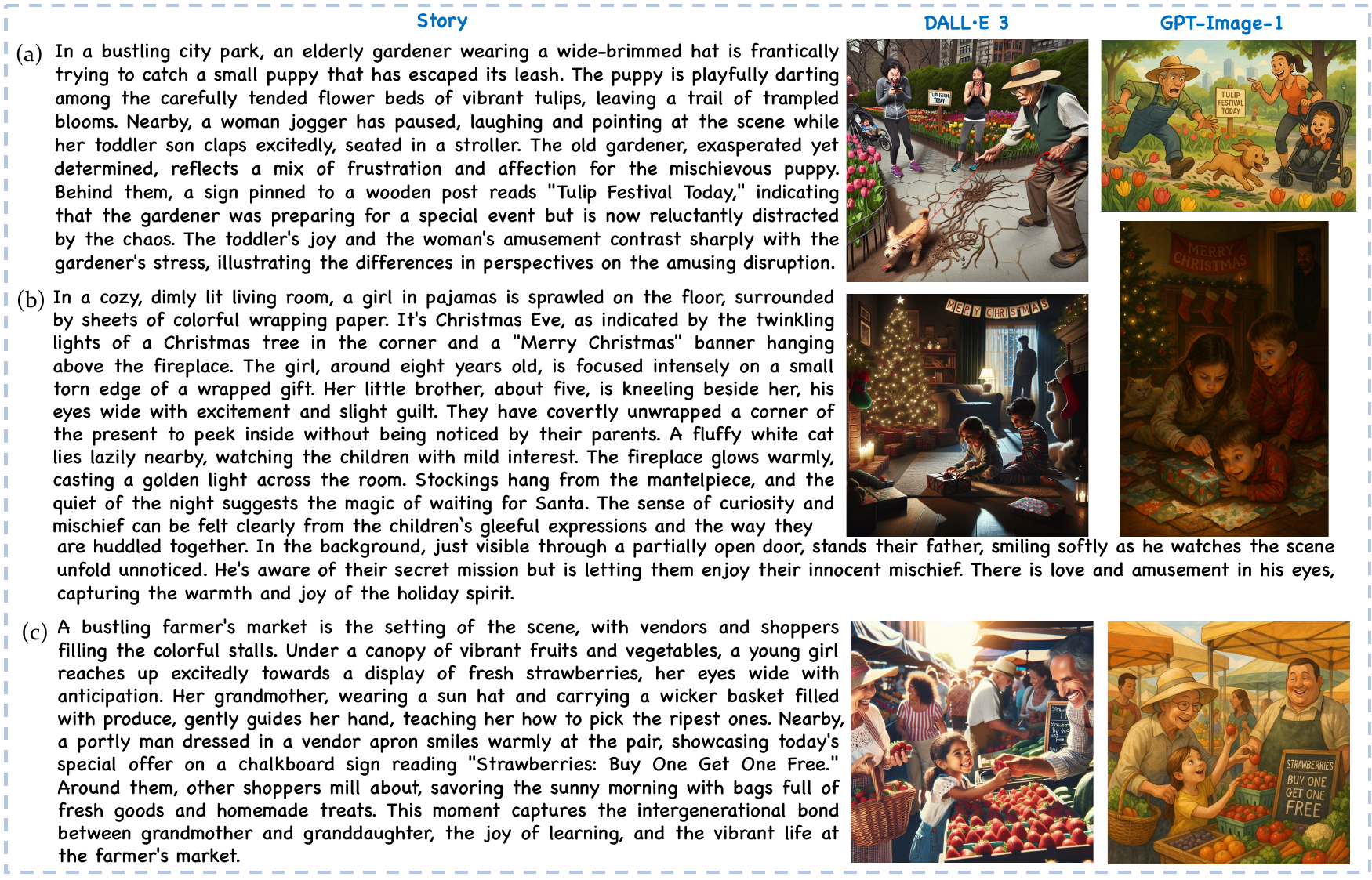}
    \caption{Comparison of generation results for the same story across different Painter models.}
    \label{fig:alignment_case}
\end{figure*}

Figure~\ref{fig:alignment_case} shows that in (a), compared to DALLE-3, GPT-image-1 more accurately and vividly depicts the text ``Tulip Festival Today'', the flowers on the ground, and people's expressions, enabling readers to more directly comprehend the story through these visual clues. 
In (b), GPT-image-1 also provides a more accurate portrayal of the characters' emotions, capturing the excitement and nervousness of the children secretly unwrapping gifts, as well as the father's sense of gratification. 
In (c), GPT-image-1 successfully renders the text “Strawberries: Buy One Get One Free” within the image, whereas DALL-E 3 fails to do so. 
Thus, we can see that the Storytelling Image Generation task also places higher demands on the capabilities of T2I models, in order to meet the requirement of telling a semantically rich story using just a single image. 

\section{Details of Evaluation Framework}
\label{sec:appendix_validations}

\subsection{Implementation and Validation of Human Semantic Score}
We adopt the annotation criteria established by~\citet{Song_Pang_Tang_Wu_Zhu_2025} to train our human evaluators and refine the scoring scale from 1 to 5 to include half-point increments for greater granularity. 
Each image is independently rated by three evaluators, and the final Human Semantic Score is obtained by averaging their scores. 
The annotators are undergraduate and graduate students aged 18 to 25. 

To measure the inter-rater reliability, we compute the Intra-class Correlation Coefficient (ICC) for our annotations. 
The model we select is two-way mixed-effects model. 
The ICC of our human evaluation is 0.88, which means there is a high degree of consistency among our evaluators. 

\subsection{Validation of KNN-based Diversity Evaluator}

\begin{table}[h!]
    \centering 
    \scriptsize 

    \begin{tabular}{lcc} 
    \hline 
    Method & Spearman & Pearson  \\
    \hline
    Avg. Distance  & 0.838 (0.009) &  0.851 (0.007) \\  
    KNN Distance & 0.802 (0.017) & 0.842 (0.009) \\  
    \hdashline
    Avg. Distance (w/ summarizer) & 0.850 (0.007) & 0.910 (0.002) \\ 
    KNN Distance (w/ summarizer) & 0.958 (0.000) & 0.912 (0.002) \\ 
    \hline
    \end{tabular}
    \caption{
    \label{tab:eval_diversity_metric}
    Validating the effectiveness of Diversity Evaluator. 
    }
\end{table}

To validate the effectiveness of our Diversity Evaluator, we first constructed a test set comprising 8 groups of data with varying levels of repetition, each containing 30 samples, and obtained human ratings for the diversity of each group. 
We then correlate our proposed diversity metric scores against these human evaluations, with the results shown in Table~\ref{tab:eval_diversity_metric}, which compares metrics using either KNN or simple Average (Avg.) distance, applied with or without our summarizer. 
According to the result, our proposed approach, KNN Distance combined with the summarizer, achieves the highest correlation with human judgments, yielding an exceptional Spearman coefficient of 0.958 and a strong Pearson coefficient of 0.912, which significantly outperform the ablations. 
Notably, the same KNN metric without the summarizer achieves only 0.802 and 0.842, highlighting the critical role of the summarizer in extracting relevant elements. 
Furthermore, the superiority of KNN over Avg. Distance confirms its effectiveness. 
These strong correlations validate that our metric robustly captures the perceived diversity of the generated stories.

\subsection{Validation of Alignment Evaluator}

To validate our Alignment Evaluator, we constructed a test set by sampling 209 key points and manually annotating their ground-truth alignment scores. 
When benchmarked against this human-annotated set, our evaluator achieved an overall accuracy of 74\%. 
This result indicates that the evaluator possesses a relatively strong ability to accurately judge whether a given keypoint is visually represented in the image.

\subsection{Usage Discussion}

In this paper, we use the evaluators to assess the effectiveness of the StorytellingPainter pipeline. 
In real-world application scenarios, combined with human supervision, the evaluators can also serve as Verifiers to assist in automatically filtering high-quality stories and images. 

\begin{figure*}[h!]
    \centering
    \includegraphics[width=0.96\textwidth]{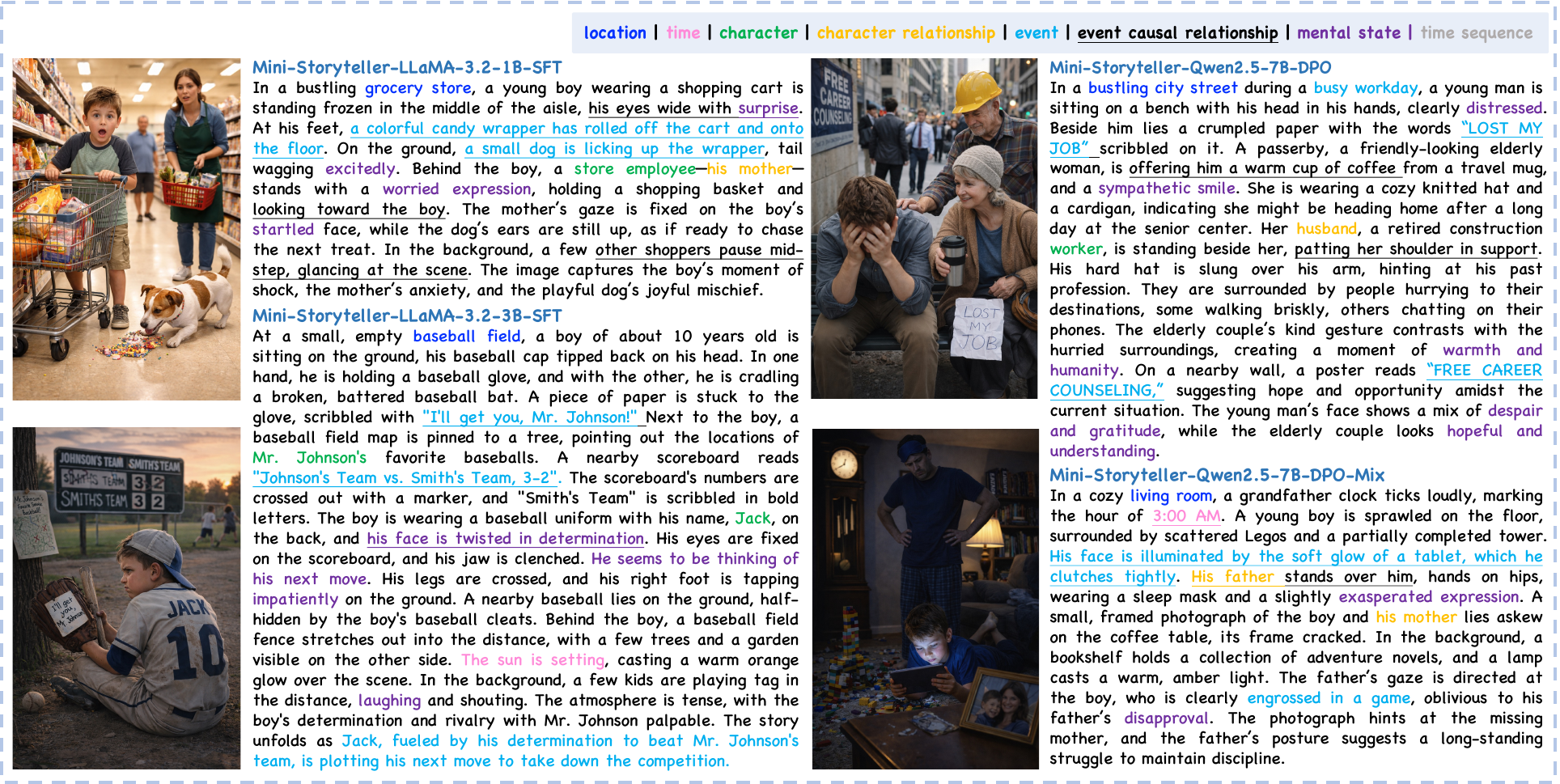}
    \caption{Stories and corresponding images generated by the StorytellingPainter pipeline using Mini-Storyteller as the Storyteller model. 
    }
    \label{fig:minis_case}
\end{figure*}

\section{Loss Functions of Mini-Storytellers}
\label{sec:loss}

\subsection{SFT}
The SFT loss function is the standard cross-entropy loss, calculated as the negative log-likelihood of the target tokens:

{\small
\begin{equation}
\mathcal{L}(\theta) = - \sum{(x, y) \in D} \sum_{t=1}^{T} \log P_{\theta}(y_t | x, y_{<t})
\end{equation}
}

where $D$ is our SFT dataset, $(x, y)$ is a prompt-response pair, $T$ is the number of tokens in the response $y$, and $P_{\theta}$ is the probability of the $t$-th token $y_t$ given the prompt $x$ and preceding tokens $y_{<t}$, as predicted by the model with parameters $\theta$.

\subsection{DPO}
The loss function used in DPO training is defined as:

{\small
\begin{align}
\mathcal{L}(\pi_\theta; \pi_{\mathrm{ref}})
&= -\mathbb{E}_{(x, y_w, y_l) \sim \mathcal{D}} \Bigg[
    \log \sigma\Bigg(
        \beta \log \frac{\pi_\theta(y_w \mid x)}{\pi_{\mathrm{ref}}(y_w \mid x)} \nonumber \\
&\qquad\qquad\qquad - \beta \log \frac{\pi_\theta(y_l \mid x)}{\pi_{\mathrm{ref}}(y_l \mid x)}
    \Bigg)
\Bigg].
\end{align}
}

Here, $\pi_\theta(y_w \mid x)$ and $\pi_\theta(y_l \mid x)$ denote the probabilities assigned by the model (parameterized by $\theta$) to two candidate responses given an input $x$, where $y_w$ is the preferred response and $y_l$ is the less preferred one. Similarly, $\pi_{\text{ref}}(y_w \mid x)$ and $\pi_{\text{ref}}(y_l \mid x)$ represent the probabilities assigned by a reference policy.

\section{Case Analysis of Mini-Storytellers}
\label{sec:case_mini}

Figure~\ref{fig:minis_case} illustrates four cases generated by four different Mini-Storytellers.
We can see that Mini-Storyteller-LLaMA-3.2-1B-SFT and Mini-Storyteller-LLaMA-3.2-3B-SFT can successfully generate concise and compelling stories without time sequence and dialogue after training. 
The images generated by Mini-Storytellers clearly contain more visual clues to form CoRs across various semantic dimensions. 
For instance, the image generated by Mini-Storyteller-LLaMA-3.2-3B-SFT conveys a vivid story of a boy determined to seek ``revenge'' after a defeat, utilizing visual clues such as the scoreboard, the boy's attire and expression, and the note. 
The image generated by Mini-Storyteller-Qwen2.5-7B-DPO depicts an unemployed man seeking counseling on the street and receiving comfort from strangers. 
For the image generated by Mini-Storyteller-Qwen2.5-7B-DPO-Mix, the sleep mask on the man's head suggests that it is already bedtime, and that he is either preparing to sleep or has just been woken up by the boy.

\section{AI Use Declaration}

Large Language Models, including Gemini-2.5-pro and GPT-5, were adopted to refine the writing in this work. 
All AI-generated text was carefully reviewed and revised by the authors.
The use of these models was limited to language refinement and did not involve any core research or analytical activities.

\end{document}